\documentclass[journal]{IEEEtran}
\usepackage{cite}
\usepackage{amsmath,amssymb,amsfonts}
\usepackage{algorithm}
\usepackage{graphicx}
\usepackage{textcomp}
\usepackage{color}
\usepackage{algpseudocode}
\usepackage[caption=false,font=footnotesize]{subfig}
\def\BibTeX{{\rm B\kern-.05em{\sc i\kern-.025em b}\kern-.08em
    T\kern-.1667em\lower.7ex\hbox{E}\kern-.125emX}}

\begin{document}

\title{Sylvas: Synergistic Learning Value based Device Scheduling in Federated Continual Learning}
\author{Yuxuan Sun,~\IEEEmembership{Member,~IEEE,} Yuxuan Bai, Tan Chen,~\IEEEmembership{Graduate Student Member, IEEE} \\Sheng Zhou,~\IEEEmembership{Senior Member,~IEEE,}
Zhisheng Niu,~\IEEEmembership{Fellow,~IEEE} 
\thanks{Y. Sun, Y. Bai are with the School of Electronic and Information Engineering, Beijing Jiaotong University, Beijing 100044, China. (e-mail:\{yxsun, yuxuanbai919\}@bjtu.edu.cn)}
\thanks{T. Chen, S. Zhou (Corresponding Author), Z. Niu are with the Department of
Electronic Engineering, Tsinghua University, Beijing 100084, China, and the Beijing National Research Center for Information Science and Technology (e-mail: chent17@tsinghua.org.cn,
\{sheng.zhou,niuzhs\}@tsinghua.edu.cn).}
}

\maketitle

\begin{abstract}
Federated continual learning (FCL) enables shared global models to continuously adapt to distributed and non-stationary data streams, making it important for Internet of things applications such as intelligent transportation, industrial monitoring, and unmanned systems. Under spatio-temporal data distribution dynamics and label scarcity, a key challenge is how to quantify the contribution of each edge device to global learning performance and schedule the most valuable devices under resource constraints for timely model updating. This article presents Sylvas, a synergistic learning value based device scheduling framework for FCL at the wireless edge. Sylvas evaluates the learning value of distributed data from two perspectives: distributional value, which characterizes the contribution of device data to global model learning from a spatio-temporal distribution perspective, and label value, which captures the quantity and reliability tradeoff of pseudo-labeled data. By integrating these factors into a synergistic learning value metric, Sylvas schedules devices with high learning value while satisfying communication and computation resource constraints. Case studies demonstrate that Sylvas supports timely model adaptation under spatio-temporal distribution dynamics and effectively exploits unlabeled data.

\end{abstract}


\begin{figure*}[!ht]
\centerline{\includegraphics[width=0.9\linewidth]{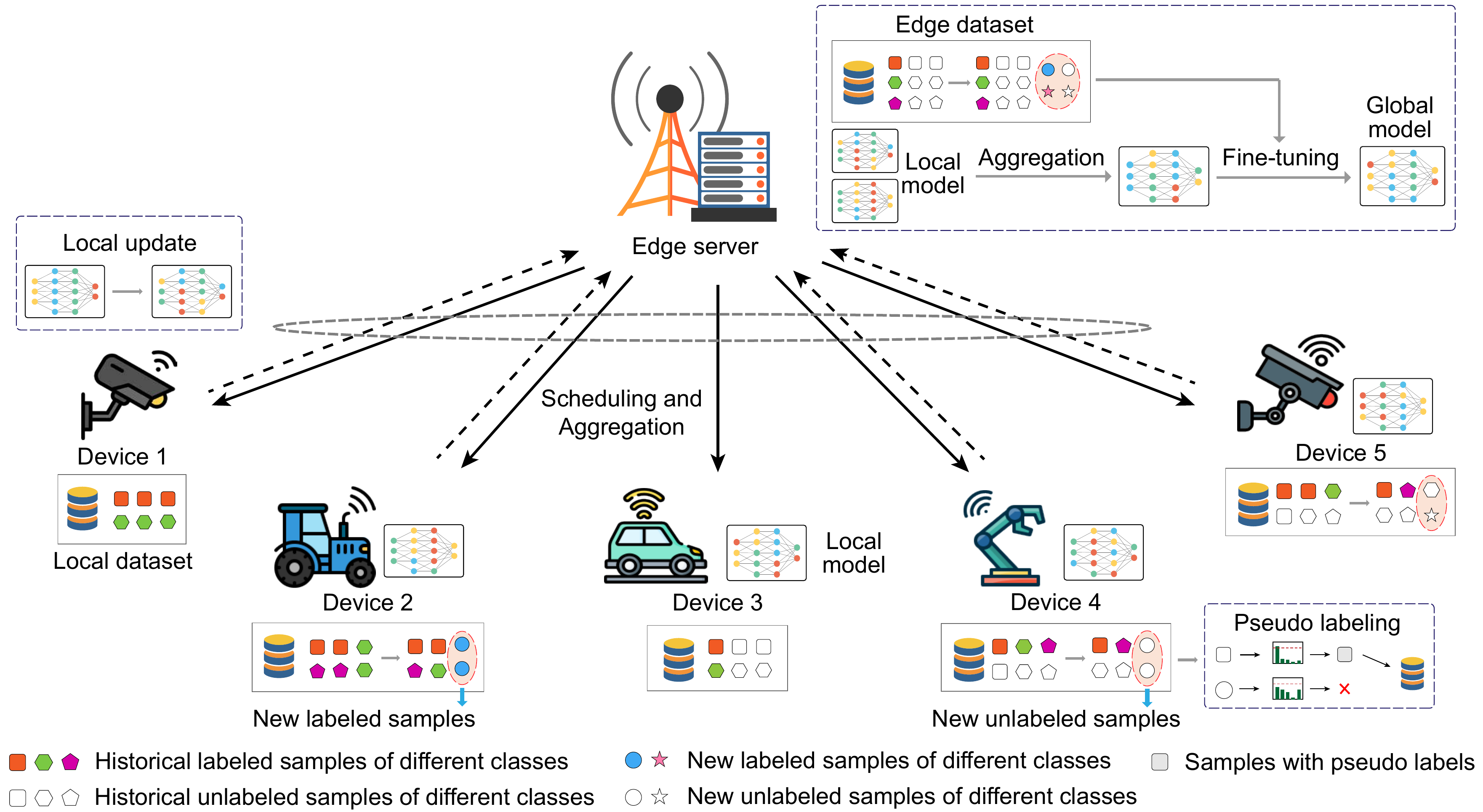}}
\caption{Illustration of federated continual learning at the wireless edge.}
\label{fig1-sys}
\end{figure*}

\section{Introduction}

The convergence of machine learning and wireless networking has accelerated the development of Internet of things and connected intelligence~\cite{dritsas2025fliot}. Smart devices, such as sensors, industrial terminals, connected vehicles, and mobile robots, continuously sense their physical environments and generate data streams for intelligent model training. These models enable a wide range of applications, including environmental perception, anomaly detection, industrial monitoring, and autonomous decision-making.

Traditional centralized learning requires raw data to be uploaded to a central server, potentially incurring substantial communication overhead, service latency, and privacy risks. Federated learning (FL) provides a promising framework for distributed model training, in which devices train models using their local data and share only model updates for global aggregation~\cite{niknam2020flwireless,tao2025fededge}. In many practical edge environments, however, data are generated continuously rather than being collected once, and their distributions evolve over time. The global model must therefore adapt to evolving data streams while retaining knowledge learned from historical data, motivating \emph{federated continual learning} (FCL)~\cite{hamedi2025fcl}. For example, connected vehicles can collaboratively update a shared perception model as road, weather, and traffic conditions evolve, while retaining knowledge of previously encountered conditions.

In this context, \emph{timely model updating} is a key objective of FCL~\cite{sun2020timelyedge}, aiming to maximize learning performance within latency budgets. Delayed updates may cause the global model to lag behind the current environment. Meanwhile, overemphasizing newly arrived data may degrade performance on previously learned data and lead to catastrophic forgetting~\cite{wang2025forgetting}. FCL therefore needs to optimize the overall learning performance across both newly arrived and historical data distributions.

Implementing FCL at the wireless edge is challenging due to limited communication bandwidth and computation capability. It is impractical to involve all devices in each training round, making device scheduling essential for timely model updating. The key question is then how to \emph{quantify the learning value} of each device, as its contribution to global learning performance is jointly affected by multiple factors. First, under spatio-temporal distribution dynamics, data distributions differ across devices and evolve over time. The learning value should therefore reflect both how much current data deviate from historical data and how well the scheduled devices collectively represent the global data distribution.
Second, continuously arriving data are often unlabeled. Although pseudo-labeling provides a practical way to exploit such data, incorrect pseudo-labels may degrade learning performance and accumulate errors over successive training rounds~\cite{uludag2025review}. Quantifying the learning value of unlabeled data is an open question.

In this article, we propose Sylvas, a \underline{sy}nergistic \underline{l}earning \underline{va}lue based device \underline{s}cheduling framework for FCL under resource constraints and label scarcity. Sylvas quantifies device learning value from two perspectives: \emph{distributional value} and \emph{label value}. Distributional value combines temporal drift, which captures distribution changes between newly arrived and historical data, with collective divergence, which measures how well the scheduled devices collectively represent the global data distribution. For unlabeled data, label value captures the tradeoff between the quantity and reliability of pseudo-labels for newly arrived and historical samples. Sylvas integrates these two perspectives into a \emph{synergistic learning value} and schedules high value devices under communication and computation constraints to support timely model updating. When all device data are labeled, Sylvas reduces to a special case that considers only distributional value. Through case studies, we demonstrate that Sylvas enables timely model adaptation under spatio-temporal distribution dynamics and effectively exploits unlabeled data in semi-supervised FCL.

In the following, we first introduce the system model and challenges of FCL in Section \ref{Sec-model}. We then present the Sylvas framework in Section \ref{Sec-Sylvas}. Through case studies in Section \ref{CaseI}, we evaluate Sylvas under fully labeled and semi-supervised FCL settings, respectively. Finally, we conclude the article and discuss future research directions in Section \ref{Conclusion}.

\section{FCL at the Wireless Edge: System Model and Device Scheduling Challenges}
\label{Sec-model}

As shown in Fig. \ref{fig1-sys}, devices such as vehicles, robots, and sensors continuously collect streaming data and use newly arriving samples to update their local models at the wireless edge. The local data distribution of each device may evolve over time as its operating state and surrounding environment change. Such evolution may result in a class-incremental setting, where previously unseen classes emerge over time. For example, autonomous vehicles may encounter new types of road users or traffic signs during operation. It may also result in a domain-incremental setting, where the label space remains unchanged but the data distribution shifts across environments or operating conditions, such as changes in weather, illumination, or road conditions. Local model updates are aggregated under the coordination of an edge server to jointly train a shared global model. Depending on annotation availability, newly collected samples may be labeled or unlabeled, giving rise to a \emph{semi-supervised} learning setting in which most device-side samples are unlabeled and the edge server maintains only a small labeled reference dataset for model initialization, calibration, and performance evaluation.

The objective of FCL is timely model adaptation, which can be formulated as maximizing learning performance within a given latency budget or, equivalently, minimizing the adaptation latency required to reach a target performance level. Unlike conventional FL, the target learning performance of FCL should jointly reflect adaptation to new data and retention of knowledge learned from historical data. In each training round, the per-round latency is determined primarily by the local computation and model transmission delays of the slowest scheduled device. Therefore, the overall adaptation latency depends jointly on the number of rounds required to reach the target performance and the per-round latency. 

Spatio-temporal distribution dynamics, label scarcity, and limited edge resources jointly complicate timely model adaptation and make device scheduling a key component of FCL at the wireless edge. Since it is impractical to involve all devices in each round, the system must identify and schedule a subset of devices with high contribution to the learning performance while satisfying resource constraints.

\subsection{Spatio-Temporal Distribution Dynamics}
Local data distributions vary across devices and continuously evolve over time. From the temporal perspective, newly arriving data may increasingly deviate from historical data, resulting in temporal data drift that reflects the emergence of new knowledge to be incorporated into the global model. From the spatial perspective, the learning contribution of a device depends not only on its own data distribution, but also on how its data complement those of other scheduled devices and how well the selected devices collectively represent the global data distribution. These two dimensions jointly determine the learning contribution of device data, motivating a \emph{distributional value} metric that captures spatio-temporal distribution dynamics for device scheduling.

\subsection{Label Scarcity}
Ground-truth labels remain scarce in dynamic edge networks. The limited labeled reference data available may not adequately represent evolving distributions or newly emerging classes. Semi-supervised strategies such as pseudo-labeling can leverage unlabeled data, but their effectiveness depends heavily on the reliability of model predictions, particularly for newly arrived or underrepresented samples. Incorrect pseudo-labels may be reinforced during local training and propagated through aggregation, causing confirmation bias and error accumulation. 
Meanwhile, the quantity of reliable pseudo-labeled samples also affects the learning performance. The fundamental tradeoff between pseudo-label quantity and reliability motivates a \emph{label value} metric that captures the learning contribution of unlabeled data for device scheduling.

\subsection{Resource Constrained Device Scheduling}

Distributed devices operate with limited and heterogeneous wireless and computation resources. Differences in channel conditions, available bandwidth, and local computing capability lead to varying communication and computation delays across devices. Under synchronous aggregation, a training round cannot complete until all scheduled devices finish their local computation and model transmission, so high-latency devices may substantially increase the per-round delay. These constraints create a tradeoff between learning value and adaptation latency. Scheduling more devices may accelerate model convergence but increases delay, whereas selecting only fast devices may reduce delay but limit learning performance. Device scheduling should balance the learning value of selected devices against their communication and computation costs.

\begin{figure*}[t]
    \centering
    \includegraphics[width=0.9\linewidth]{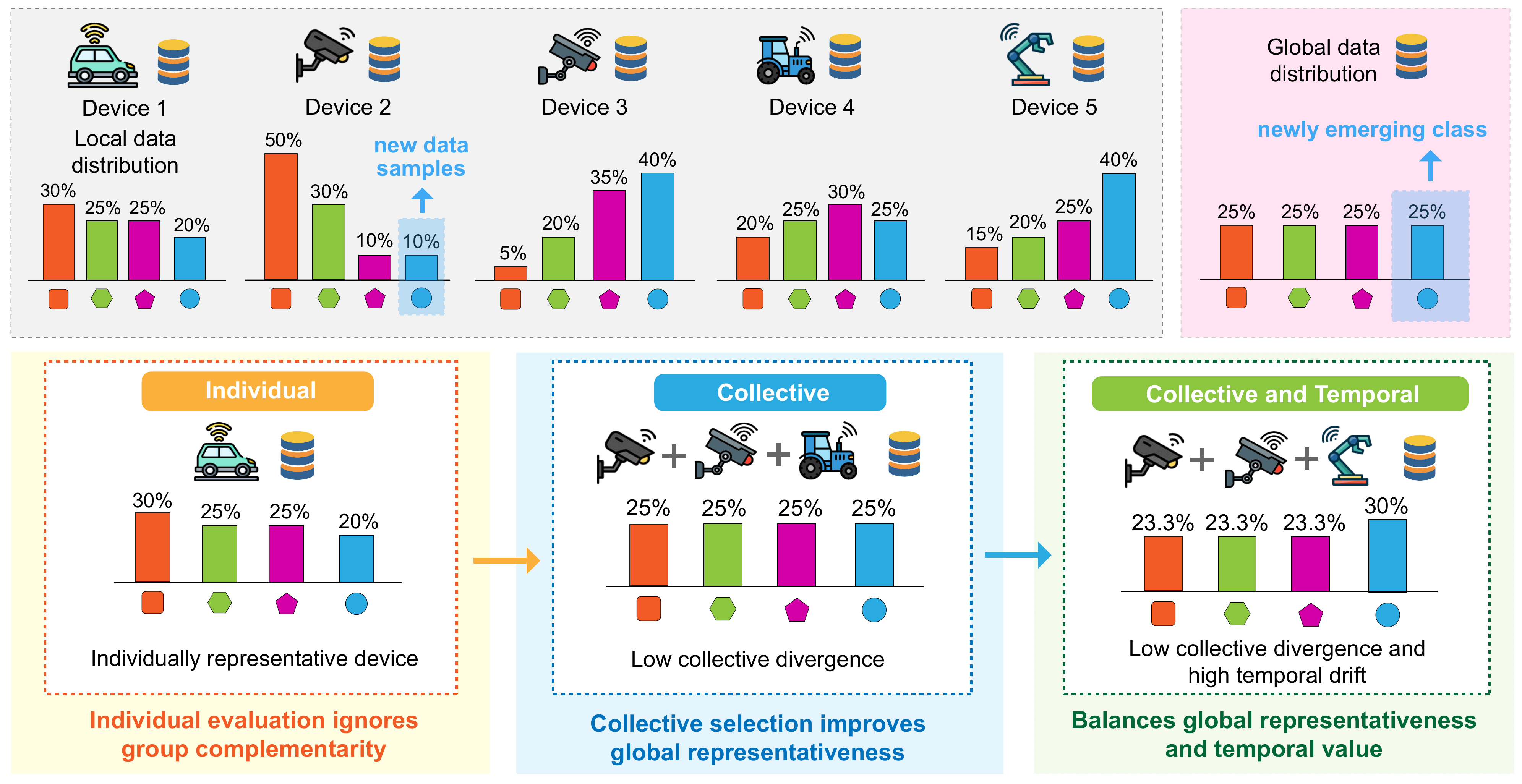}
    \caption{Conceptual illustration of device scheduling under spatio-temporal distribution dynamics. Sylvas jointly captures device-level temporal drift and group-level collective divergence to characterize the distributional value of each device.}
    \label{fig:FCL_workflow}
\end{figure*}

\section{Sylvas: Synergistic Learning Value based Device Scheduling}
\label{Sec-Sylvas}

To address the above challenges, we propose Sylvas, a device scheduling framework for FCL that evaluates device learning value under resource constraints. The key idea of Sylvas is to characterize device learning value synergistically from two perspectives: distributional value and label value. Distributional value captures the contribution of device data under spatio-temporal distribution dynamics, while label value captures the learning contribution of unlabeled data by balancing pseudo-label quantity and reliability. These learning-value metrics are integrated into a \emph{synergistic learning value}, which is then combined with device computing capabilities and wireless communication conditions to schedule high-value devices under resource and latency constraints.

\subsection{Distributional Value for Spatio-Temporal Dynamics}

As illustrated in Fig.~\ref{fig:FCL_workflow}, the distributional value in Sylvas characterizes the learning contribution of device data by combining collective divergence, which measures how well the scheduled devices collectively represent the global data distribution, with temporal drift, which captures distribution changes between newly arrived and historical data. 

In conventional FL, device scheduling typically prioritizes devices with smaller gradient divergence, defined as the discrepancy between the local and global gradients, while scheduling as many devices as possible under resource constraints. Our recent study shows that data heterogeneity is better characterized as a collective property of the scheduled device group rather than an independent property of each device~\cite{fedcgd}. Even when individual devices have biased local data distributions, combining devices with complementary data can make their aggregate data distribution closer to the global distribution, thereby accelerating model convergence. Based on this observation, we introduce \emph{collective divergence} to quantify the discrepancy between the aggregate data distribution of the scheduled devices and the global data distribution. It is defined in the gradient space as the discrepancy between the weighted average gradient of the scheduled device group and the gradient over the global dataset. For classification tasks, this quantity can be further upper bounded by the weighted discrepancy between the aggregate class distribution of the scheduled devices and the global class distribution, where the distributional discrepancy of each class is weighted by its corresponding class-specific gradient norm.

For temporal distribution evolution, the scheduler should prioritize devices whose newly arrived data exhibit substantial distributional changes, while retaining sufficient historical data to mitigate catastrophic forgetting. Accordingly, Sylvas adopts replay-based continual learning, where each device trains on newly arrived data together with a limited set of representative historical samples. We introduce \emph{temporal drift} to quantify the discrepancy between current and historical data distributions of each device~\cite{fedteddi}. It is defined in the gradient space as the discrepancy between the gradients induced by the current and historical local data of each device. For classification tasks, this quantity can be further characterized according to the form of data evolution. In a class-incremental setting, temporal drift can be upper bounded by the weighted changes in class proportions between consecutive stages, where the historical proportion of a newly emerging class is set to zero. In a domain-incremental setting, the class proportions remain unchanged, and temporal drift can be characterized by the differences in the average model updates induced by the same classes across consecutive stages. 

Collective divergence and temporal drift jointly form the distributional value. Sylvas employs a time-varying scheduling objective that rewards high temporal drift while penalizing high collective divergence, thereby balancing rapid adaptation with global representativeness. Early in training, Sylvas places greater weight on temporal drift to accelerate adaptation to newly arrived data. As training proceeds, it gradually shifts the weight toward collective divergence, improving the representativeness of the selected device group and stabilizing performance across newly arrived and historical data.

\begin{figure*}[t]
    \centering
    \includegraphics[width=1.0\linewidth]{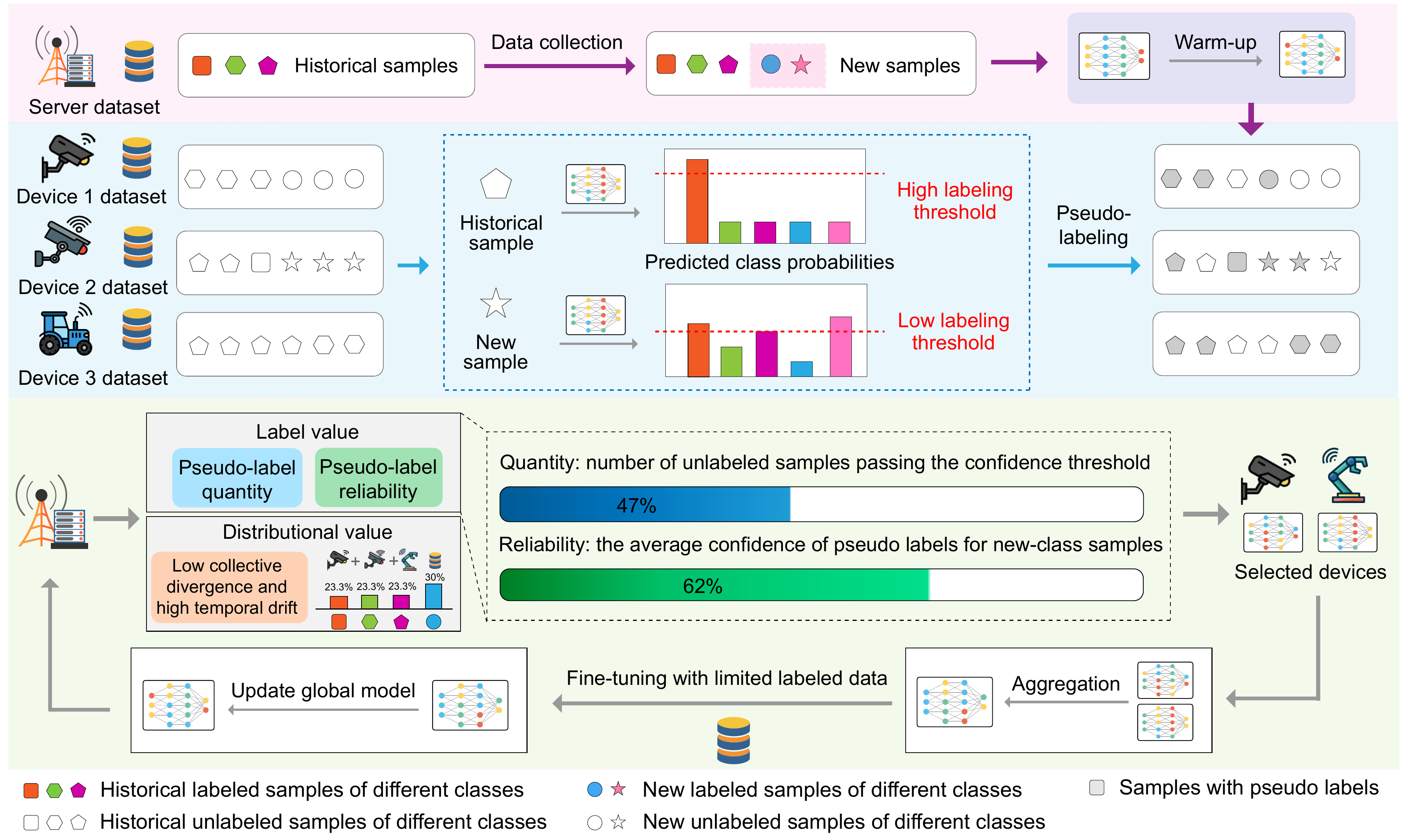}
    \caption{Workflow of Sylvas for semi-supervised FCL, where label value balances pseudo-label quantity and reliability.}
    \label{fig:FSSL_workflow}
\end{figure*}

\subsection{Label Value for Unlabeled Data}

To effectively exploit unlabeled data, Sylvas adopts a semi-supervised learning mechanism that generates and filters pseudo-labels for local samples and evaluates their \emph{label value} based on pseudo-label quantity and reliability. The workflow is illustrated in Fig.~\ref{fig:FSSL_workflow}. We consider a setting in which the edge server maintains a small labeled reference dataset, while most device-side data remain unlabeled. The edge server first initializes the global model using the labeled reference data and then broadcasts it to the distributed devices. In each training round, each candidate device uses the received global model to generate pseudo-labels for its local unlabeled samples and applies confidence-threshold-based filtering to retain only samples whose pseudo-label confidence exceeds a predefined threshold for local training. After local training, the server aggregates the received model updates and further fine-tunes or calibrates the global model using the labeled reference data.

Label value characterizes the fundamental tradeoff between \emph{pseudo-label quantity and reliability}. Specifically, pseudo-label quantity is the number of unlabeled samples passing the confidence threshold, while pseudo-label reliability is estimated by the average confidence of the retained pseudo-labels for each class. A larger pseudo-label quantity allows more samples to participate in local training and provides richer information for model updating, whereas higher pseudo-label reliability reduces the adverse effect of incorrect pseudo-labels and improves learning accuracy.

Moreover, newly introduced and previously learned classes often exhibit different pseudo-labeling characteristics. For newly introduced classes, the model has not yet learned sufficiently discriminative representations. An overly high confidence threshold may therefore filter out many potentially valuable samples and slow adaptation to new knowledge. Therefore, Sylvas initially adopts a relatively relaxed filtering criterion for newly introduced classes to retain sufficient effective pseudo-labeled samples. As the model progressively acquires more discriminative representations and its predictions become more reliable, the confidence threshold is gradually increased to suppress incorrect pseudo-labels and prevent error accumulation. In contrast, predictions for previously learned classes are generally more stable, allowing a stricter filtering criterion to be applied throughout training. This \emph{adaptive thresholding strategy} enables Sylvas to accelerate early adaptation while progressively improving pseudo-label reliability.

\subsection{Sylvas under Resource Constraints}

Sylvas integrates distributional value and label value into a \emph{synergistic learning value} to characterize the overall contribution of each candidate device group to model updating. Under resource constraints, however, a device group with high synergistic learning value may also incur high training latency. In synchronous FCL, the per-round latency is determined by the local computation and uplink communication delays of the scheduled devices, with the slowest device potentially dominating the round completion time. This creates a tradeoff between learning gain and latency: selecting more high-value devices may improve the accuracy of the global model and reduce the number of rounds to convergence, but may also increase per-round latency.


Sylvas therefore performs resource-constrained device scheduling by jointly considering synergistic learning value and device-side computation and communication costs. In each training round, the edge server selects a feasible device group that provides high synergistic learning value while satisfying the prescribed latency and resource constraints. Since collective divergence depends on the overall composition of the selected device group, this selection cannot be achieved by independently ranking individual devices, resulting in a challenging combinatorial optimization problem. Sylvas provides a fixed-sum coordinate descent strategy that iteratively exchanges scheduled and unscheduled devices for each candidate group size and selects the best resulting device group. To further reduce computational complexity, Sylvas also adopts a greedy strategy that iteratively selects the device providing the largest marginal gain in synergistic learning value without violating the latency and resource constraints. 

\section{Case Studies}
\label{CaseI}
In this section, we evaluate Sylvas under two representative FCL settings. The first case considers fully labeled data, where Sylvas operates as a special case based only on distributional value to address spatio-temporal distribution dynamics. The second case extends Sylvas to semi-supervised FCL, where label value is further incorporated to exploit device-side unlabeled data through pseudo-label quantity and reliability.

\subsection{Sylvas with Fully Labeled Data}
We first evaluate Sylvas in a fully labeled class-incremental FCL system under latency constraints. This setting represents a special case of Sylvas that considers only distributional value. We use the CIFAR-100 dataset to evaluate the continual learning performance, and consider an FCL system with 30 devices and one edge server. ResNet-18 is adopted as the learning model. The model is initially trained on class indexes 0--29, and new classes are then introduced gradually, after which classes 30--39 and 40--44 are introduced in successive stages. At each incremental stage, 12 devices receive samples from the newly introduced classes. The per-round latency budget is set to 30s.

Sylvas schedules devices using a time-varying combination of temporal drift and collective divergence. Specifically, the weight assigned to temporal drift is initialized to 2.0 at the beginning of each incremental stage and decreases linearly to zero over the communication rounds. Therefore, Sylvas places greater emphasis on temporal drift in the early rounds to prioritize devices containing newly arrived classes. As training proceeds, the temporal drift weight gradually decreases, shifting the scheduling emphasis toward collective divergence to improve the global representativeness of the selected device group and stabilize model performance.

For comparison, we consider seven baselines. Random scheduling selects devices randomly in each round. Best Channel prioritizes devices with favorable wireless channel conditions~\cite{amiri2021updateaware}. Power-of-Choice selects devices with larger local losses, while Best Norm gives priority to devices with larger local update norms~\cite{cho2022powerofchoice}. FedCBS addresses spatial data heterogeneity by selecting devices based on class imbalance in their local datasets~\cite{fedcbs}. Our previous work, FedCGD, schedules device groups solely according to collective divergence~\cite{fedcgd}. We also include Pure Drift as an ablation baseline that selects devices solely based on temporal drift.

\begin{figure}[t]
\centering
\includegraphics[width=0.95\linewidth]{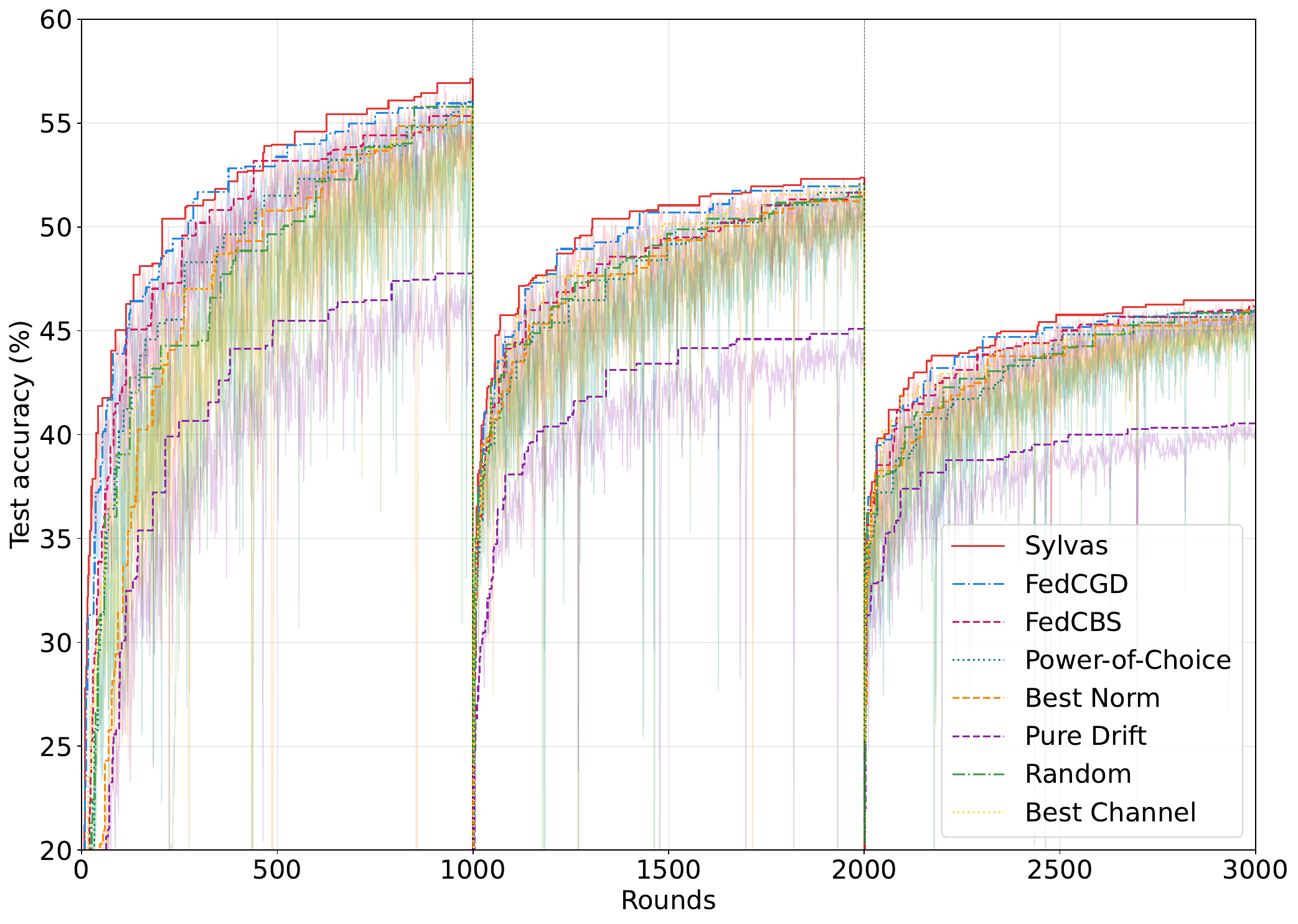}
\caption{Performance of Sylvas with fully labeled data under spatio-temporal distribution dynamics.}
\label{fig:case1}
\end{figure}

Fig.~\ref{fig:case1} shows the test accuracy of different scheduling schemes during the continual learning process. The lightly shaded curves represent the round-by-round test accuracy, while the solid curves indicate the best test accuracy achieved up to each communication round. The arrival of new classes causes a temporary accuracy drop for all methods, reflecting the need to adapt to the changed data distribution. Random scheduling and Best Channel cannot fully capture the learning value of newly arrived data, while FedCBS and FedCGD mainly emphasize distributional representativeness under non-i.i.d. data. Pure Drift can quickly respond to data changes, but may over-emphasize new data and increase the risk of forgetting. By jointly considering temporal drift and collective divergence, Sylvas selects device groups that capture newly emerging knowledge while maintaining global representativeness. Consequently, it recovers more rapidly after each distribution change and maintains more stable performance across continual learning stages.

\subsection{Sylvas in Semi-Supervised FCL}
\label{CaseII}

We further evaluate Sylvas in a latency-constrained semi-supervised FCL system. The objective is to exploit device-side unlabeled data under limited server-side annotations while avoiding unreliable pseudo-labels that may mislead continual model updates. The edge server first initializes the global model using its limited labeled data and broadcasts the model to the devices. Candidate devices generate pseudo-labels for their local unlabeled samples and report lightweight statistics describing pseudo-label quantity and reliability. Based on these statistics, the server then schedules a subset of devices for local training. Only the scheduled devices train on the retained pseudo-labeled samples and upload their model updates. After aggregation, the server further fine-tunes the global model using its labeled reference data.

The system consists of 30 devices and one edge server. A convolutional neural network is trained on the SVHN dataset. The dataset contains 10 classes, with class indexes 0--5 used as the initial classes. At the incremental stage, the edge server maintains 1,000 labeled samples from previously learned classes and 100 labeled samples from newly introduced classes. Each device holds 800 unlabeled samples, while only 12 devices contain data from the newly introduced classes. The per-round latency budget is set to 1s. 
For pseudo-label generation, Sylvas prioritizes newly introduced classes by gradually increasing their confidence threshold from 0.5 to 0.8 as training proceeds, while fixing the threshold for previously learned classes at 0.95. Samples that satisfy neither threshold are excluded from the unsupervised loss.

\begin{figure}[t]
\centering
\subfloat[Communication rounds required to reach different target accuracies.]{
\includegraphics[width=0.95\linewidth]{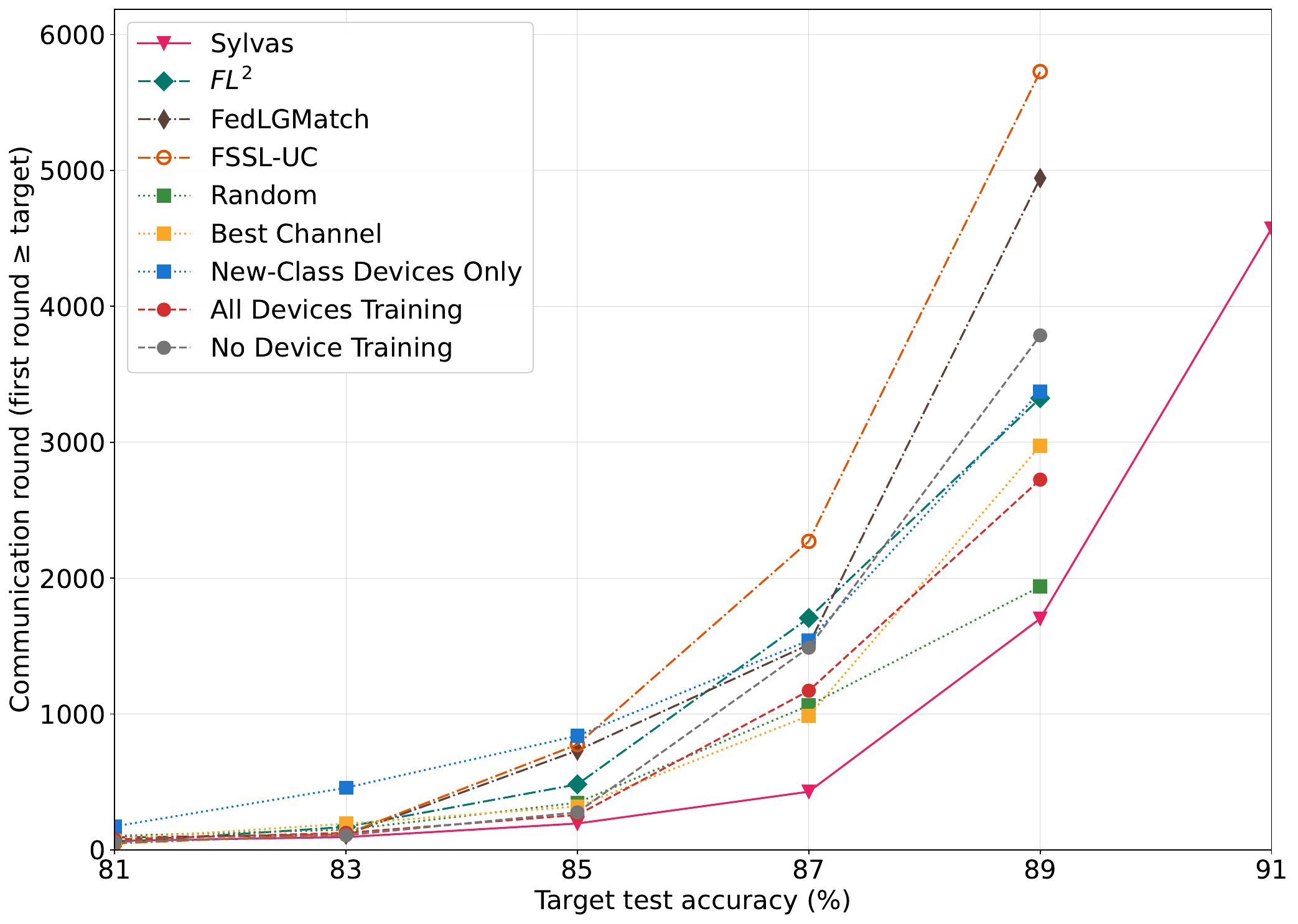}
\label{fig:case2a}
}
\vspace{0.5em}
\subfloat[Test accuracy over communication rounds.]{
\includegraphics[width=0.95\linewidth]{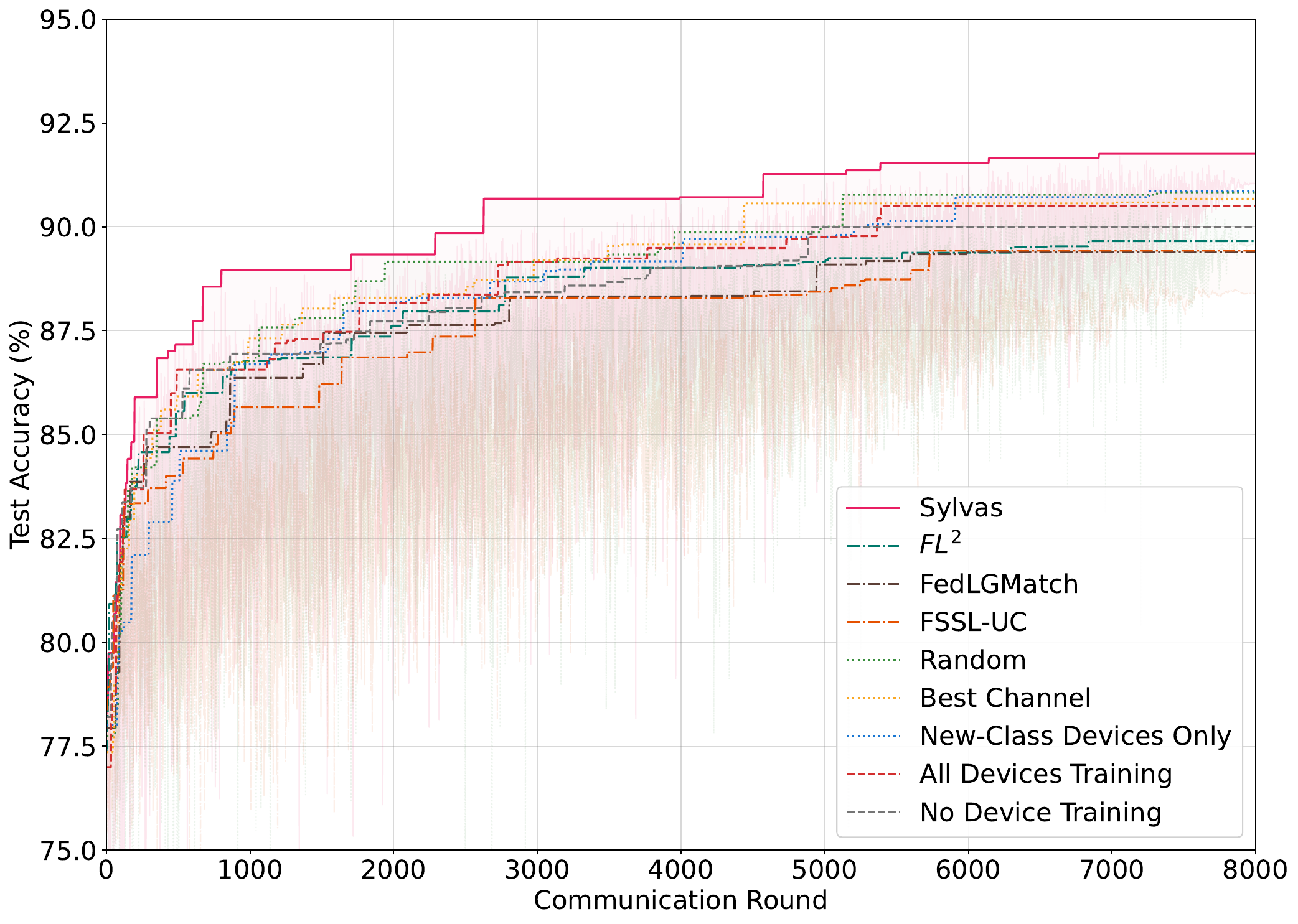}
\label{fig:case2b}
}
\caption{Performance of Sylvas in semi-supervised FCL.}
\label{fig:case2}
\end{figure}

For comparison, we consider the following baselines. New-Class Devices Only selects only devices containing newly arrived classes, while All Devices Training lets all available devices participate in training. No Device Training only uses the limited labeled data at the edge server without involving device-side unlabeled data. We further compare Sylvas with representative federated semi-supervised learning methods, including $\text{FL}^2$ for few-label learning~\cite{fl2}, FedLGMatch for local-global pseudo-label matching~\cite{fedlgmatch}, and FSSL-UC for uncertainty and consistency aware training~\cite{fssluc}.

Fig.~\ref{fig:case2} shows the performance of different methods under massive unlabeled data. As shown in Fig.~\ref{fig:case2}(a), Sylvas requires fewer communication rounds to reach different target accuracies, with a more pronounced advantage at higher accuracy targets. This indicates that simply using more unlabeled data, selecting only devices with new classes, or relying only on favorable channel conditions is not sufficient for continual learning with limited labeled data. Sylvas instead prioritizes devices that offer a favorable balance between pseudo-label quantity and reliability. As shown in Fig.~\ref{fig:case2}(b), Sylvas achieves higher test accuracy during continual training, indicating that it can better exploit unlabeled data and improve model generalization under resource constraints. As all methods operate under the same per-round latency budget, the reduction in communication rounds directly translates into lower adaptation latency. Overall, these results demonstrate that balancing pseudo-label quantity and reliability improves model performance with limited labeled data and the timeliness of FCL.

\section{Conclusion and Outlook}
\label{Conclusion}

This article presented Sylvas, a synergistic learning value based device scheduling framework for FCL at the wireless edge. The key challenge addressed by Sylvas is how to quantify device learning value when data distributions are spatially heterogeneous and temporally evolving, and large volume of data remain unlabeled. Sylvas addresses this challenge from two perspectives. Distributional value combines collective divergence and temporal drift to characterize the value of device data to global model learning from a spatio-temporal distribution perspective. Label value captures the tradeoff between pseudo-label quantity and reliability to characterize the learning contribution of unlabeled data. These two metrics are integrated into a synergistic learning value and further combined with communication and computation costs for resource-constrained device scheduling. The case studies under fully labeled and semi-supervised FCL settings have shown that Sylvas improves learning performance under evolving data distributions, reduces the communication rounds required for model adaptation,  and enables effective use of unlabeled data.

Several future directions deserve further investigation. First, mobility-aware FCL is important for dynamic edge systems with vehicles, drones, and mobile robots, where device mobility changes both wireless channels and local data distributions. The interplay among mobility, data drift, and communication opportunities should be jointly modeled. Second, active learning can be integrated with FCL. Since labels are scarce and expensive, the system may actively select the most valuable or uncertain samples for annotation, thereby reducing labeling costs and improving new-class learning. Third, resource-efficient continual learning requires effective training-triggering mechanisms. Instead of updating models at fixed intervals, future systems may trigger training only when significant data drift or performance degradation is detected. Designing reliable drift detection and training-triggering policies under privacy and resource constraints remains an important open problem.


\begin{IEEEbiographynophoto}
{Yuxuan Sun} (yxsun@bjtu.edu.cn) is an associate professor in the School of Electronic and Information Engineering, Beijing Jiaotong University, China. Her research interests include edge intelligence, federated learning, task-oriented communications and vehicular networks.
\end{IEEEbiographynophoto}
\vspace{-.5in}

\begin{IEEEbiographynophoto}
{Yuxuan Bai} (yuxuanbai919@bjtu.edu.cn) is a master's student in the School of Electronic and Information Engineering, Beijing Jiaotong University. Her major research interests include federated learning, continual learning, and semi-supervised learning.
\end{IEEEbiographynophoto}
\vspace{-.5in}

\begin{IEEEbiographynophoto}
{Tan Chen} (chent17@tsinghua.org.cn) is a PhD student in the Department of Electronic Engineering, Tsinghua University. His major research interests include federated learning and edge intelligence.
\end{IEEEbiographynophoto}
\vspace{-.5in}

\begin{IEEEbiographynophoto}
{Sheng Zhou} (sheng.zhou@tsinghua.edu.cn) is an associate professor in the Department of Electronic Engineering, Tsinghua University. His research interests include vehicular networks, mobile edge computing, and green wireless communications.
\end{IEEEbiographynophoto}
\vspace{-.5in}

\begin{IEEEbiographynophoto} 
{Zhisheng Niu} (niuzhs@tsinghua.edu.cn) is a professor in the Department of Electronic Engineering, Tsinghua University. His major research interests include queueing theory, traffic engineering, radio resource management of wireless networks, and green communication and networks.
\end{IEEEbiographynophoto}

\end{document}